\documentclass{ecai}

\pdfoutput=1
\usepackage{times}
\usepackage{graphicx}
\usepackage{latexsym}
\graphicspath{{img/}}
\usepackage{amsmath}
\usepackage{booktabs}
\usepackage{amsfonts}

\usepackage[T1]{fontenc}
\usepackage[utf8]{inputenc}

\usepackage{hyperref}
\usepackage{fancyhdr}
\usepackage{lipsum}
\usepackage[utf8]{inputenc}
\usepackage{algorithm}
\usepackage[noend]{algpseudocode}
\usepackage{multirow}
\usepackage{float}
\usepackage{array}

\algblock{Input}{EndInput}
\algnotext{EndInput}
\algblock{Output}{EndOutput}
\algnotext{EndOutput}
\usepackage[normalem]{ulem}
\usepackage{color}
\usepackage{subfigure}

\begin{document}

\title{Generating Natural Language Adversarial Examples on a Large Scale with Generative Models}




\author{Yankun Ren\institute{Ant Financial Services Group, Emails: \{yankun.ryk, jianbin.ljb, jun.zhoujun, shuang.yang, yuan.qi\}@anfin.com}~~,
~~Jianbin Lin$^1$~~,
~~Siliang Tang$^{*,}$\institute{Zhejiang University, Email: siliang@zju.edu.cn}~~,
~~Jun Zhou$^1$~~,
~~Shuang Yang$^1$~~,
~~Yuan Qi$^1$ 
\and Xiang Ren\institute{University of Southern California, Email: xiangren@usc.edu \newline* Corresponding author.}}


\maketitle

\begin{abstract}
Today text classification models have been widely used. However, these classifiers are found to be easily fooled by adversarial examples.
Fortunately, standard attacking methods generate  adversarial texts in a pair-wise way, that is, an adversarial text can only be created from a real-world text by replacing a few words. In many applications, these texts are  limited in numbers, therefore their corresponding adversarial examples are often not diverse enough and sometimes hard to read, thus can be easily detected by humans and cannot create chaos at a large scale.
In this paper, we propose an end to end solution to efficiently generate adversarial texts from scratch using generative models, which are not
restricted to perturbing the given texts. We call it unrestricted adversarial text generation. Specifically, we train a conditional variational autoencoder (VAE) with an additional adversarial loss to guide the generation of adversarial examples. Moreover, to improve the validity of adversarial texts, we utilize discrimators and the training framework of generative adversarial networks (GANs) to make adversarial texts consistent with real data.
Experimental results on sentiment analysis demonstrate the scalability and efficiency of our method. It can attack text classification models with a higher success rate than existing methods, and provide acceptable quality for humans in the meantime. 
\end{abstract}

\section{Introduction}
 
Today machine learning classifiers have been widely used to  provide key services such as information filtering, sentiment analysis. 
However, recently researchers have found that these ML classifiers, even deep learning classifiers are vulnerable to adversarial attacks. They demonstrate that image classifier~\cite{goodfellow2014explaining} and now even text classifier~\cite{ren2019generating}  can be fooled easily by adversarial examples that are deliberately crafted by attacking algorithms.
Their algorithms generate adversarial examples in a pair-wise way. That is,
given one input $x \in \mathcal{X}$, they aim to generate one corresponding adversarial example ${x}'\in \mathcal{X}$ by adding small imperceptible perturbations to $x$. 
The adversarial examples must maintain the semantics of the original inputs, that is, ${x}'$ must be still classified as the same class as $x$ by humans. 
On the other hand, adversarial training is shown to be a useful defense method to resist adversarial examples~\cite{szegedy2013intriguing, goodfellow2014explaining}. Trained on a mixture of adversarial and clean examples, classifiers can be resistant to adversarial examples.

In the area of natural language processing (NLP), existing methods are pair-wise, thus heavily depend on input data $x$. If attackers want to generate adversarial texts which should be classified as a chosen class with pair-wise methods, they must first collect texts labeled as the chosen class, then transform these labeled texts to the corresponding adversarial examples by replacing a few words. As the amount of labeled data is always small, the number of generated adversarial examples is limited. These adversarial examples are often not diverse enough and sometimes hard to read, thus can be easily detected by humans. Moreover, in practice, if attackers aim to attack a public opinion monitoring system, they must collect a large number of high-quality labeled samples to generate a vast amount of adversarial examples, otherwise, they can hardly create an impact on the targeted system. Therefore, pair-wise methods only demonstrate the feasibility of the attack but cannot create chaos on a large scale.

\begin{figure}[!t]
\centering
\includegraphics[width=0.45\textwidth, angle=0]{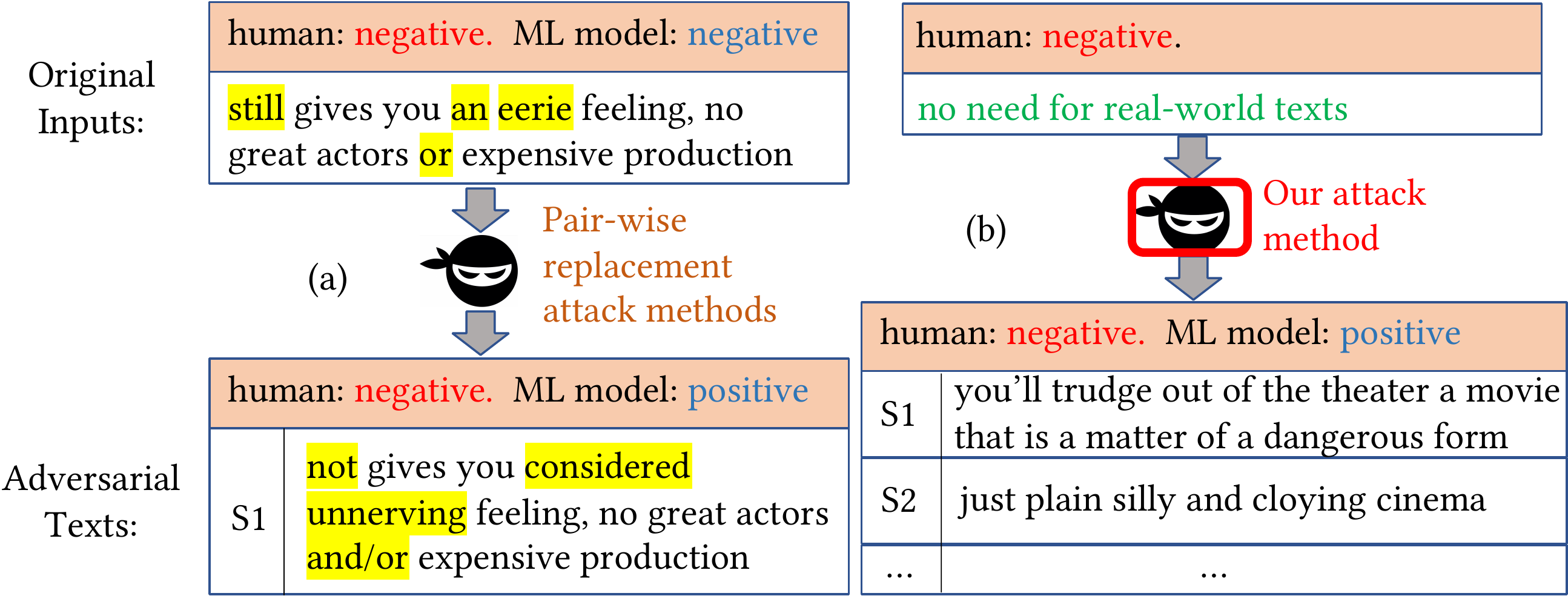}
\caption{An illustration of adversarial text generation.  (a) Given one negative text which is also classified as negative by a ML model,  traditional methods replace a few words (yellow background) in the original text to get one paired adversarial text, which is still negative for humans, but the model prediction changes to positive. (b) Our unrestricted method does not need input texts. We only assign a ground-truth class - negative, then our method can generate large-scale adversarial texts. which are negative for humans, but classified as positive by the ML model.}
\label{fig:motivation}
\end{figure}

In this paper, we propose an unrestricted end to end solution to efficiently generate adversarial texts, where adversarial examples can be generated from scratch without real-world texts and are still meaningful for humans. We argue that adversarial examples do not need to be generated by perturbing existing inputs. For example, we can generate a movie review that does not stem from any examples in the dataset at hand. If the movie review is thought to be a positive review by humans but classified as a negative review by the targeted model, the movie review is also an adversarial example. Adversarial examples generated in this way can break the limit of input number, thus we can get large scale adversarial examples. On the other hand, the proposed method can  also be used to create more adversarial examples for defense. Trained with more adversarial examples often means more robustness for these key services. 

The proposed method leverages a conditional variational autoencoder (VAE) to be the generator which can generate texts of a desired class. To guide the generator to generate texts that mislead the targeted model, we access the  targeted model in a white-box setting and use an adversarial loss to make the targeted model make a wrong prediction. In order to make the generated texts consistent with human cognition, we use discrimators and the training framework of generative adversarial networks (GANs) to make generated texts similar as real data of the desired class.  After the whole model is trained, we can sample from the latent space of VAE and generate infinite adversarial examples without accessing the targeted model. The model can also transforms a given input to an adversarial one.

We evaluate the performance of our attack method on a sentiment analysis task. Experiments show the scalability of generation. The adversarial examples generated from scratch achieve a high attack success rate and have acceptable quality. As the model can generate texts only with feed-forwards in parallel, the generation speed is quite fast compared with other methods. Additional ablation studies verify the effectiveness of discrimators, and data augmentation experiments demonstrate that our method can generate large-scale adversarial examples with higher quality than other methods. When existing data at hand is limited, our method is superior over the pair-wise generation.

In summary, the major contributions of this paper are as follows:
\begin{itemize}
    \item Unlike the existing literature in text attacks, we aim to construct adversarial examples not by transforming given texts. Instead, we train a model to generate text adversarial examples from scratch. In this way, adversarial examples are not restricted to existing inputs at hand but can be generated from scratch on a  large-scale. 
    \item We propose a novel method based on the vanilla conditional VAE. To generate adversarial examples,  we incorporate an adversarial loss to guide the vanilla VAE's generation process.
    \item We adopt one discrimator for each class of data. When training, we train the discrimators and the conditional VAE in a min-max game like GANs, which can make generated texts more consistent with   real data of the desired class.
    \item We conduct attack experiments on a sentiment analysis task. Experimental results show that our method is scalable and achieves a higher attack success rate at a higher speed than recent baselines. The quality of generated texts is also acceptable. Further ablation studies and data augmentation experiments verify our intuitions and demonstrate the superiority of scalable text adversarial example generation.
\end{itemize}

\section{Related Work}
There has been extensive studies on adversarial machine leaning, especially on deep neural models~\cite{szegedy2013intriguing, goodfellow2014explaining, liang2018deep, samanta2017towards, alzantot2018generating}.
Much   work  focuses on image classification tasks \cite{szegedy2013intriguing,goodfellow2014explaining, carlini2017towards,hu2017generating,xiao2018spatially}.  \cite{szegedy2013intriguing} solves the attack problem as an optimization problem with a box-constrained L-BFGS.   \cite{goodfellow2014explaining} proposes the fast gradient sign method (FGSM), which perturbs images with noise computed as the gradients of the inputs.

In NLP, perturbing texts is more difficult than images, because words in sentences are discrete,  on which we can not directly perform gradient-based attacks like continuous image space.
Most methods adapt  the pair-wise methods of image attacks  to text attacks. They perturb texts by replacing a few words in texts. 
\cite{papernot2016crafting, gong2018adversarial, cheng2018seq2sick} calculate gradients with respect to the word vectors and perturb word embedding vectors with gradients. They find the  word  vector nearest to the perturbed vector. In this way, the perturbed vector can be map to a discrete word to replace the original one. These methods are gradient-based replacement methods.  

Other attacks on texts can be summarized as gradient-free replacement methods. They replace words in texts  with typos or synonyms. 
\cite{liang2018deep} proposes to edit words with tricks like insertion, deletion and replacement. They choose appropriate words to replace by calculating the word frequency and the highest gradient magnitude. 
\cite{li2019textbugger} proposes five automatic word replacement methods, and use magnitude of gradients of the word embedding vectors to choose the most important words to replace.  \cite{ren2019generating} is based on synonyms substitution strategy. Authors introduce a new word replacement order determined by both the word saliency and the classification probability. However, these  replacement methods still generate adversarial texts in a pair-wise way, which restrict the adversarial texts to the variants of given real-world texts. Besides, the substitute words sometimes change text meanings. Thus existing adversarial text generation methods only demonstrate the feasibility of the attack but cannot create chaos on a large scale. 

In order to tackle the  above problems, we propose an unrestricted end to end solution to generate diverse adversarial texts on a large scale with no need of given texts. 


\section{Methodology}
In this section, we propose a novel method to generate adversarial texts for the text classification model on a large scale. Though trained with labeled data in a pair-wise way, after it is trained,  our model can generate an unlimited number of adversarial examples without any input data. Moreover, like other traditional pair-wise generation methods, our model can also transform a given text into an adversarial one. Unlike the existing methods, our model generates adversarial texts without querying the attacked model, thus the generation procedure is quite fast.

\begin{figure}[!t]
\centering
\includegraphics[width=0.45\textwidth, angle=0]{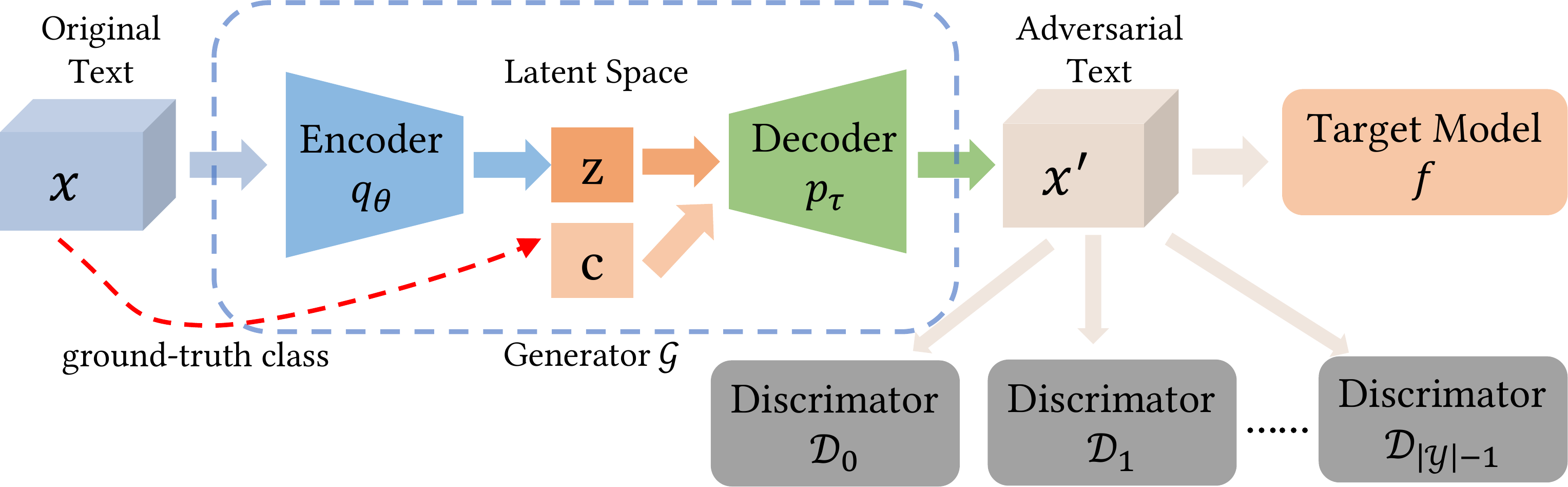}
\caption{The architecture of the whole model. In the training phase, $\mathcal{G}$ generates an adversarial text ${x}'$ to reconstruct the original text $x$, and feed ${x}'$ to $\mathcal{D}$ and $f$ to make $f$ predict differently on $x$ and ${x}'$. After trained, the model can generate large-scale adversarial texts based on sampled latent space vector $z$ and a chosen class $c$ without original texts $x$.}
\label{fig:whole_model}
\end{figure}

\subsection{Overview}

Figure \ref{fig:whole_model} illustrates the overall architecture of our model. The model has three components: a generator $\mathcal{G}$,  discrimators $\mathcal{D}$, and a targeted model $f$. $\mathcal{G}$ and $\mathcal{D}$ form a generative adversarial network (GAN). When training, we feed an original input  to the generator $\mathcal{G}$, which transforms $x$ to an adversarial output ${x}'$. The procedure can be defined as follows:
\begin{equation}
    \mathcal{G}(x):x\in \mathcal{X}\rightarrow {x}'
    \label{eq:g}
\end{equation}

$\mathcal{G}$ aims to generate ${x}'$ to reconstruct $x$. Then, we feed the generated ${x}'$ to the targeted model $f$, and $f$ will classify  ${x}'$ as a certain class, which we hope is a wrong label. Thus we have the following equation:
\begin{equation}
    f({x}')=y_{t} \in \mathcal{Y}
\end{equation}
where $y_t \neq f(x)$ and $\mathcal{Y}$ is the label space of the targeted classification model.

In order to keep ${x}'$ being classified as the same class as $x$ by human, we add one discrimator for each class $y \in \mathcal{Y}$. With the help of the min-max training strategy of GAN framework, each class $y$'s discrimator can make ${x}'$ close to the distribution of real class $y$ data, thus  ${x}'$ is made to be compatible with human congnition.

We now proceed by introducing these components in further details.

\subsection{Generator} \label{section:text_generator}

In this subsection,  we describe the generator $\mathcal{G}$ for text generation. We use the variational autoencoder (VAE) \cite{kingma2013auto, rezende2014stochastic} as the generator. The VAE is a generative model based on a regularized version of the standard autoencoder. This model supposes the latent variable $z$ is sampled from a prior distribution.

As shown in Figure \ref{fig:whole_model}, the VAE is composed of the encoder $q_{\theta }(z|x)$ and the decoder $p_{\tau }(x|z)$, where $\tau$ is the parameters of $p$ and $\theta$ is the parameters of $q$. $q_{\theta}$ is a neural network. Its input is a text $x$, its output is a latent code $z$. $q_{\theta}$ encodes $x$ into a latent representation space $\mathcal{Z}$, which is a lower-dimensional space than the input space.  
$p_{\tau }$ is another neural network. Its input is the code $z$, it outputs an adversarial text ${x}'$ to the probability distribution of the input data $x$.

\begin{figure}[!t]
\centering
\includegraphics[width=0.47\textwidth, angle=0]{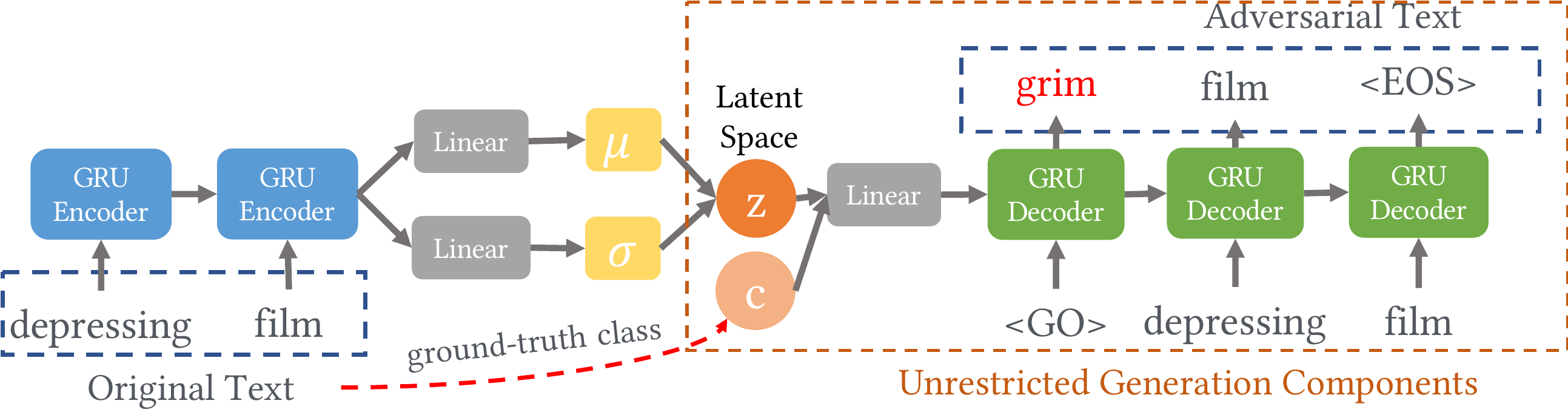}
\caption{The generator $\mathcal{G}$. When training, we need input texts to train $\mathcal{G}$ After $\mathcal{G}$ is trained, we only need to sample $z$ from the latent space, and use the decoder to generate adversarial texts unrestrictedly without original texts.}
\label{fig:generator}
\end{figure}

In our model, we adopt the gated recurrent unit (GRU) \cite{cho2014properties} as the encoder and the decoder. As in Figure \ref{fig:generator}, The input $x$ is a sentence of words, we formulate the input for neural networks as follows: for a word at the position $i$ in a sentence, we first transform it into a word vector $v_i$ by looking up a word embedding table. The word embedding table is randomly initialized and is updated during the model training. Then the word embedding vectors are fed into the GRU encoder. In the $i$-th GRU cell, a hidden state $h_i$ is emitted.

We use $h_N$ to denote the last GRU cell's hidden state, where N is the length of the encoder input. In order to get latent code $z$, we feed $h_N$ into two linear layers to get $\mu$ and $\sigma$ respectively. Following the Gaussian reparameterization trick \cite{kingma2013auto}, we sample a random sample $\varepsilon $ from a standard Gaussian ($\mu = \overrightarrow{0}$, $\sigma = \overrightarrow{1}$), and compute $z$ as:
\begin{equation}
    z = \mu + \sigma \circ \varepsilon 
\end{equation}
Computed in this way, $z$ is guaranteed to be sampled from a Gaussian distribution $\mathcal{N}(\mu, \sigma^2)$.

Then, we can decode $z$ to generate an adversarial text ${x}'$. Before feeding $z$ to the decoder, we adopt a condition embedding $c_k$ to guide the decoder to generate text ${x}'$ of a certain class $y_k$, which can be chosen arbitrarily. Suppose in a text classification task, there are $\left | \mathcal{Y} \right |$ classes.  Specifically, we randomly initialize a class embedding table as a matrix $C\in R^{\left | \mathcal{Y} \right | \times d}$ and look up $C$ to get the corresponding embedding $c_k$ of class $y_k$. Then, we feed $[z, c_k]$ into a linear layer to get another vector representation. The vector encodes the information of the input text and a desired class.


The decoder GRU uses this vector as the initial state to generate the output text. Each GRU cell generates one word. The computation process is similar to that of the GRU encoder, except the output layer of each cell. The output $\mathbf{O}_i$ of the $i$-th GRU cell is computed as:
\begin{equation}
    \mathbf{u}_i =  W_h \cdot h_i
    \label{eq:cell_out}
\end{equation}
\begin{equation}
    \mathbf{O}_{i, k} =  \frac{e^{\mathbf{u}_{i, w_k}}}{\sum_{j=1}^{\left | \mathcal{V} \right |}e^{\mathbf{u}_{i, w_j}}}
\end{equation}
where $W_h  \in \mathbb{R}^{d_{h_i} \times  \left | \mathcal{V} \right |}$  is the transformation weights, $\mathcal{V}$ is the word vocabulary, $w_k \in \mathcal{V}$ and $u_i,  \mathbf{O}_{i} \in \mathbb{R}^{\left | \mathcal{V} \right |}$. $\mathbf{O}_{i, k}$ is the probability of the $i$-th GRU cell emitting the $k$-th word $w_k$ in the vocabulary. 

In the training phase, the GRU cell chooses the word index with the highest probability to emit:
\begin{equation}
    w_k = \mathop{\arg\max}_{k} \mathbf{O}_{i, k}
    \label{eq:argmax}
\end{equation}

When training, the loss function of the VAE is calculated as:
\begin{equation}
\begin{split}
\mathcal{L}_{VAE}(\theta , \phi )=&-\mathbb{E}_{q_{\theta}(z|x)}(\log p_{\tau }(x|z))\\
&+\alpha \mathbb{KL}(q_{\theta}(z|x) || p(z))
\end{split}
\label{eq:vae_loss}
\end{equation}

The first term is the reconstruction loss, or expected negative log-likelihood. This term encourages the decoder to learn to reconstruct the data. So the output text is made to be similar to the input text. The second term is the Kullback-Leibler divergence between the latent vector distribution $q_{\theta}(z|x)$ and $p(x)$. If the VAE were trained with only the reconstruction objective, it would learn to encode its inputs deterministically by making the variances in $q(z|x)$ vanishingly small \cite{raiko2014techniques}. Instead, the VAE uses the second term to encourages the model to keep its posterior distributions close to a prior $p(z)$, which is generally set as a standard Gaussian.

In the training phase, the input to  the GRU decoder is the input text, appended with a special $<$GO$>$ token as the start word. We add a special $<$EOS$>$ token to the input text as the ground truth of the output text. The $<$EOS$>$ token represents the end of the sentence. When training the GRU decoder to generate texts, the GRU decoder tends to ignore the latent code $z$ and only relies on the input to emit output text. It actually degenerates into a language model. This situation is called KL-vanishing. To tackle the KL-vanishing problem in training GRU decoder, we adopt the KL-annealing mechanism \cite{bowman2016generating}. KL-annealing mechanism gradually increase the KL weight $\alpha$ from $0$ to $1$. This can be thought of as annealing from a vanilla autoencoder to a VAE. Also, we randomly drop the input words into the decoder with a fixed keep rate $k \in [0, 1]$, to make the decoder depend on the latent code $z$ to generate output text. 

Notably, if we randomly sample $z$ from a standard Gaussian, the decoder can also generate output text based on $z$. The difference is that there is no input to the GRU decoder, but we can send the word generated by the $i$-th GRU cell to the $(i+1)$-th GRU cell as the $(i+1)$-th input word. Specifically,
in the inference phase, we use beam-search to generate words. The initial input word to the first GRU cell is the $<$GO$>$ token. When the decoder emits the $<$EOS$>$ token, the decoder stops generating new words, and the generation of one complete sentence is finished. 

In this way, after $\mathcal{G}$ is trained, theoretically, we can sample infinite $z$ from the latent space and generate infinite output texts based on these $z$. This is part of the superiority of our method.


\subsection{Targeted Model} \label{sec:targeted_model}
Since the TextCNN model has good performances and is quite fast, it is one of the  most widely used methods for text classification task in industrial applications \cite{zhang2019adversarial}. As we aim to attack models used in practice,  we take the TextCNN model \cite{kim2014convolutional} as our targeted model. 

Suppose we set the condition of the VAE to be $y_k$, the decoder generates the output text ${x}'$, then we feed the text into the targeted model, and the targeted model will predict a probability $P_{target}(y_i)$ for each candidate class $y_i$. We conduct targeted attack and aim to cheat the targeted model to classify ${x}'$ as class $y_t$ ($y_t \neq y_k$), we can get the following adversarial loss function:
\begin{equation}
    \mathcal{L}_{adv}=-\mathbb{E}_{p_{\tau }(x|z)}(\log P_{target}(y_t))
    \label{eq:adv_loss}
\end{equation}
This is a cross entropy loss that maximize the probability of class $y_t$.

Recall that words in the  adversarial text ${x}'$  are computed in Equation \ref{eq:argmax}, in which Function $\arg\max$ is not derivative. So we can not directly feed the word index computed in Equation \ref{eq:argmax} into the targeted model. In this paper, we utilize the Gumbel-Softmax \cite{jang2016categorical} to make continuous value approximate discrete word index. The embedding matrix $W$ fed to  TextCNN is calculated as:
\begin{equation}
    \widetilde{\mathbf{O}}_{i, k} =  \frac{\exp(\log ((\mathbf{u}_{i, w_k})+g_k)/t ) }{\sum_{j=1}^{\left | \mathcal{V} \right |}\exp(\log ((\mathbf{u}_{i, w_j})+g_k)/t )}
    \label{eq:gumbel}
\end{equation}

\begin{equation}
    W_i = \widetilde{\mathbf{O}}_{i} \cdot E
    \label{eq:gumbel_emb}
\end{equation}
where $E \in \mathbb{R}^{\left | V \right | \times d_w}$ is the whole vocabulary embedding matrix, $u_i$ is from Equation \ref{eq:cell_out}, $g_k$ is drawn from $Gumbel (0, 1)$ distribution \cite{jang2016categorical} and $t$ is the temperature.

\begin{algorithm}[!t]
\caption{Text Adversarial Examples Generation}
\label{alg:gen}
\begin{flushleft}
{\bf Input:}
Training data of different classes $\mathbf{X}_0$, ..., $\mathbf{X}_{|\mathcal{Y}|-1}$ \\
{\bf Output:}
Text Adversarial Examples 
\end{flushleft}
\begin{algorithmic}[1]
\State Train a VAE by minimizing $\mathcal{L}_{VAE}$ on $\mathbf{X}_0$, ..., $\mathbf{X}_{|\mathcal{Y}|-1}$ with KL-annealing mechanism and word drop
\State Initialize $\mathcal{G}$ with the pretrained VAE
\State Initialize the targeted model with a pretrained TextCNN
\State Freeze the weights of the targeted model
\Repeat
\For{$y_k=y_0, y_1, ..., y_{|\mathcal{Y}|-1}$}
\State sample a batch of $n$ texts $\{x_i\}_{i=0}^n$ of class $y_k$ from $\mathbf{X}_k$
\State $\mathcal{G}$ generates $\{{x}'\}_{i=0}^n$ with condition $c_k$ 
\State \parbox[t]{150pt}{ Compute $\mathcal{L}_{disc}^{k} = \frac{1}{n}\sum \limits_{i=1}^n \log \mathcal{D}_k(x)  + \frac{1}{n}\sum \limits_{i=1}^n \log (1-\mathcal{D}_k({x}'))$ \strut}
\EndFor
\State \textbf{end for}
\State \parbox[t]{200pt}{ Update weights of $\mathcal{D}_0$, $\mathcal{D}_1$, ..., $\mathcal{D}_{|\mathcal{Y}|-1}$ by minimizing $-\sum_{k=1}^{|\mathcal{Y}|-1}\mathcal{L}_{disc}^{k}$ \strut}
\State Update weights of $\mathcal{G}$  by minimizing $\mathcal{L}_{joint}$
\Until convergence
\If {With inputs for the encoder}
\State \parbox[t]{200pt}{Encode inputs and decode the corresponding adversarial texts\strut}
\Else
\State Randomly sample $z \in \mathcal{N}(0, 1)$ and choose a class $y_k \in \mathcal{Y}$
\State \parbox[t]{200pt}{ The decoder takes $[z, c_k]$ and generates the adversarial text from scratch \strut}
\EndIf
\end{algorithmic}
\end{algorithm}

\subsection{Discrimator Model}
Until this point, ideally, we suppose the generated ${x}'$ should have many same words as $x$ of class $y_k$ (thus be classified as $y_k$ by humans) and be classified as class $y_t$ by the targeted model. But this assumption is not rigorous.  Most of the time, ${x}'$ is not classified as $y_k$ by humans. In natural language texts, even a single word change may change the whole meaning of a sentence.  A valid adversarial example must be imperceptible to humans. That is, humans must classify ${x}'$ as class $y_k$. 

Suppose $X_k$ is the distribution of real data of class $y_k$ and ${X_k}'$ is the distribution of generated adversarial data transformed from $x \in X$.
We utilize the idea of GAN framework to make ${x}'$ similar to data from $X_k$. Thus ${x}'$ will be classified as $y_k$ by humans and classified as $y_t$ at the same time.

Specifically, we adopt one discrimator $\mathcal{D}_k$ for each class $y_k \in \mathcal{Y}$. $\mathcal{D}_k$ aims to distinguish the data distribution of real labeled data $x$ of class $y_k$ and adversarial data ${x}'$ generated by $\mathcal{G}$ with desired class $y_k$:
\begin{equation}
    \mathcal{L}_{disc}^{k} = \mathbb{E}_{x \sim X_k}[\log (\mathcal{D}_k(x))] + \mathbb{E}_{{x}' \sim {X_k}'}[\log (1-\mathcal{D}_k({x}'))]
\end{equation}

The overall training objective is a min-max game played between the generator $\mathcal{G}$ and the discrimators $\mathcal{L}_{disc}^{0}$, $\mathcal{L}_{disc}^{1}$, ..., $\mathcal{L}_{disc}^{|\mathcal{Y}|-1}$, where $|\mathcal{Y}|$ is the total number of classes:
\begin{equation}
    \mathop{min}\limits_{\mathcal{G}}\mathop{max}\limits_{\mathcal{D}_k}\mathcal{L}_{disc}^{k}
    \label{eq:disc_loss}
\end{equation}

$\mathcal{D}_k$ tries to distinguish $X_k$ and ${X_k}'$, while $\mathcal{G}$ tries to fool $\mathcal{D}_k$ to make ${x}' \in {X_k}'$ be classified as real data by $\mathcal{D}_k$. Trained in this adversarial way, the generated adversarial text distribution ${X_k}'$ is drawn close to distribution $X_k$, which is of class $y_k$. Thus ${x}'$ is mostly likely to be similar to data from $X_k$ and is classfied as $y_k$ by human as a result.

We implement the discrimators with multi-layer perceptions (MLPs). Because $\arg\max$ function is not derivable, similar to Equation \ref{eq:gumbel} and \ref{eq:gumbel_emb} in Section \ref{sec:targeted_model}, we first use Gumbel-Softmax to transform the decoder output $u_i$ from Equation \ref{eq:cell_out} into a fixed-sized matrix $V = [w_1, w_2, \dots , w_m]^T$. Then, $\mathcal{D}_k$ calculate the probability of a text being true data of class $y_k$ as:
\begin{equation}
    \mathcal{D}_k(x) = \textup{MLP}(V)
\end{equation}

\subsection{Model Training}

Combining Equations \ref{eq:vae_loss},  \ref{eq:adv_loss}, \ref{eq:disc_loss}, we obtain the joint loss function for model training:
\begin{equation}
    \mathcal{L}_{joint} = \mathcal{L}_{VAE} + \phi \mathcal{L}_{adv} +  \sum\limits_{k=1}^{|\mathcal{Y}|-1}\mathcal{L}_{disc}^{k} 
    \label{eq:joint_loss}
\end{equation}

We first train the VAE and the targeted model $f$ with training data. Then we freeze weights of the targeted model and initialize  the $\mathcal{G}$'s weights  with the pretrained VAE's weights. At last, the generator $\mathcal{G}$ and all the discrimators $\mathcal{L}_{disc}^{0}$, $\mathcal{L}_{disc}^{1}$, ..., $\mathcal{L}_{disc}^{|\mathcal{Y}|-1}$ are trained in a min-max game with loss $\mathcal{L}_{joint}$. The whole training process is  summarized in Algorithm \ref{alg:gen}.

\section{Experiments}

We report the performances of our method on attacking TextCNN on sentiment analysis task, which is an important text classification task. Sentiment analysis  is widely applied to helping a business understand the social sentiment of their products or services by monitoring online user reviews and comments \cite{pang2008opinion, cambria2016affective, morente2019carrying}. In several experiments,  we evaluate the quality of the text  adversarial examples for sentiment analysis generated by the proposed method.

Experiments are conducted from two aspects. 
Specifically, we first follow the popular settings and evaluate our model's performances of transforming an existing input text into an adversarial one. We observe that our method has higher attack success rate, generates fluent texts and is efficient. Besides, we also evaluate our method on generating adversarial texts from scratch unrestrictedly. Experimental results show that we can generate large-scale diverse examples. The generated adversarial texts are mostly valid, and can be utilized to substantially improve the robustness of text classification models. 

We further report ablation studies, which verifies the effectiveness of the discrimators. Defense experiment results demonstrate that generating large-scale can help to make model more robust.

\subsection{Experiment Setup and Details}
Experiments are conducted on two popular public benchmark datasets. They are both widely used  in sentiment analysis \cite{tiwari2017implementation, moh2015multi, firmanto2018prediction} and adversarial example generation \cite{li2019textbugger, sato2018interpretable, song2018fooling}.

\smallskip
\noindent
\textbf{Rotten Tomatoes Movie Reviews (RT) \cite{pang2005seeing}.} This dataset consists of $5,331$ positive and $5,331$ negative processed movie reviews. 
We divide $80\%$ of the dataset as the training set, $10\%$ as the development set and $10\%$ as the test set.

\smallskip
\noindent
\textbf{IMDB \cite{maas2011learning}.} This dataset contains 50,000 movie reviews  from online movie websites. It consists of positive and negative paragraphs.
25,000 samples are for training and 25,000 are for testing. We held out $20\%$ of the training set as a validation set as \cite{li2019textbugger}.



\subsection{Comparing With Pair-wise Methods} \label{section:pair-wise}

In most of the existing work \cite{ren2019generating, michel2019evaluation, alzantot2018generating}, text adversarial examples are generated through a pair-wise way. That is, first we should take a text example, and then  transform it into an adversarial instance. 

To compare with the current methods fairly, we limit our method to pair-wise generation. In this experiment, we set $\phi = 9$. Specifically, we first feed an input text into the GRU encoder, and set the condition $c_k$ as the ground-truth class of the text. After that, the decoder can decode $[z, c_k]$ to get the adversarial output text. 

We choose four representative methods as baselines:

\begin{itemize}
  \item \textbf{Random}: Select $10\%$ words  randomly and modify them. 
  \item \textbf{Fast Gradient Sign Method (FGSM)} \cite{goodfellow2014explaining}: 
  First, perturbation is computed as $\varepsilon$sign($\bigtriangledown_x J$), where $J$ is the loss function and $x$ is the word vectors. 
  Then, search in the word embedding table to find the nearest word vector to the perturbed word vector. 
  FGSM is the fastest among gradient-based replacement methods.
  \item \textbf{DeepFool} \cite{moosavi2016deepfool}: This is also a gradient-based replacement method.
  It aims to find out the best direction, towards which it takes the shortest distance to cross the decision boundary. The perturbation is also applied to the word vectors. After that, nearest neighbor search is used to generate adversarial texts.
  \item \textbf{TextBugger} \cite{li2019textbugger}: TextBugger is a gradient-free replacement method. 
  It proposes strategies such as changing the word's spelling and replacing a word with its synonym, to change a word slightly to create adversarial texts. Gradients are only computed to find the most important words to change.
\end{itemize}

\smallskip
\noindent
\textbf{Attack Success Rate.}
Following the existing literature \cite{goodfellow2014explaining, moosavi2016deepfool, li2019textbugger}, we evaluate the attack success rate of our method and four baseline methods.

\begin{table}
\caption{Attack success rate of transforming given texts in a pair-wise way.}
\label{table:trans_success}
\centering
\begin{tabular}{l|c|c}
\toprule
\textbf{Method}      & \textbf{RT} & \textbf{IMDB}\\ \toprule
Random               & $1.5\%$  & $1.3\%$                      \\ 
FGSM+NNS             & $25.7\%$  & $36.2\%$                     \\ 
DeepFool+NNS         & $28.5\%$  & $23.9\%$                     \\ 
TextBugger           &    $85.1\%$  & $90.5\%$                     \\ 
Ours ($\phi = 5$) & $\mathbf{87.1\%}$  & $\mathbf{92.8\%}$                    \\ \bottomrule
\end{tabular}
\end{table}

\begin{figure*}[!t]
\centering
\includegraphics[width=0.9\textwidth, angle=0]{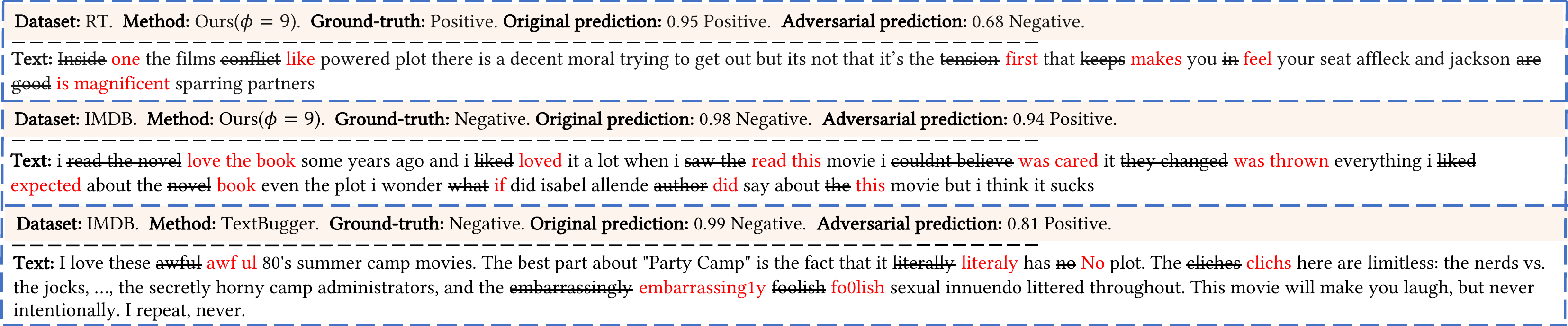}
\caption{Adversarial texts generated in a pair-wise way. In texts, the crossed out contents are from the original texts, while the red texts are the substitute contents in the adversarial examples.}
\label{fig:trans_examples}
\end{figure*}

\begin{figure*}[!t]
\centering
\subfigure[Attack Success Rate]{
\label{fig:random_gen.suc_rate}
\includegraphics[width=0.28\textwidth]{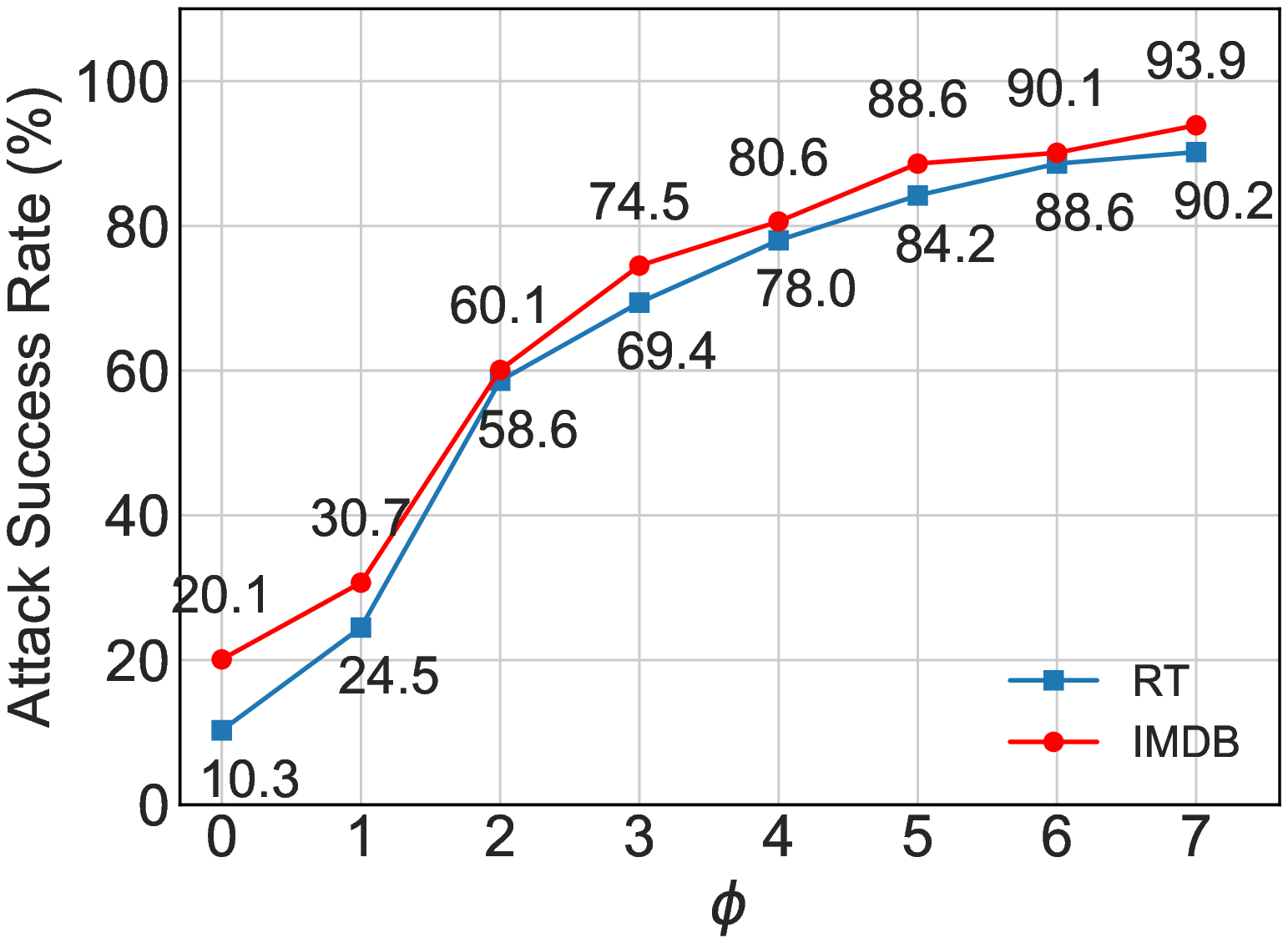}
}
\subfigure[Perplexity]{
\label{fig:random_gen.perplexity}
\includegraphics[width=0.28\textwidth]{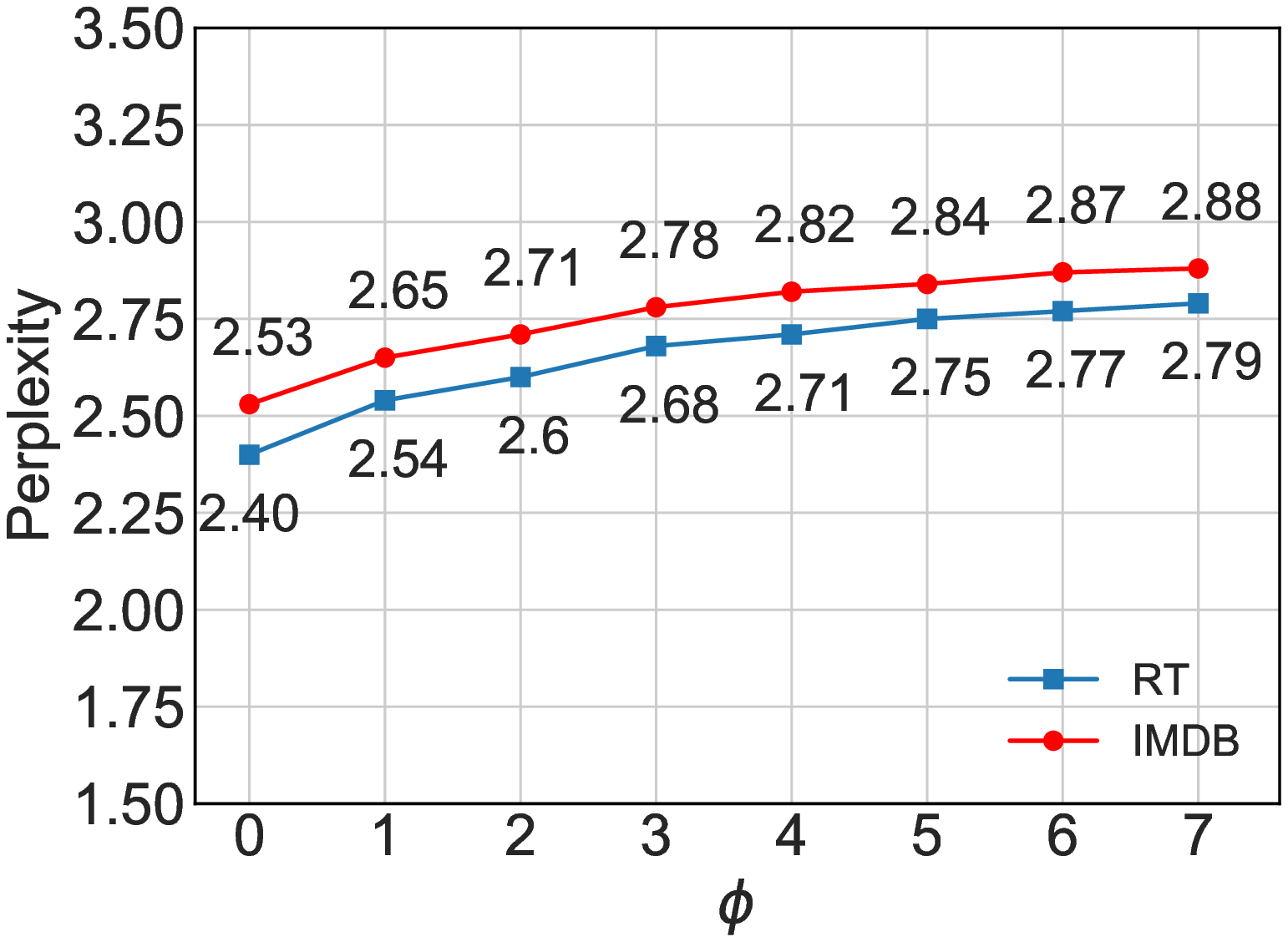}
}%
\subfigure[Validity Rate]{
\label{fig:random_gen.validity}
\includegraphics[width=0.28\textwidth]{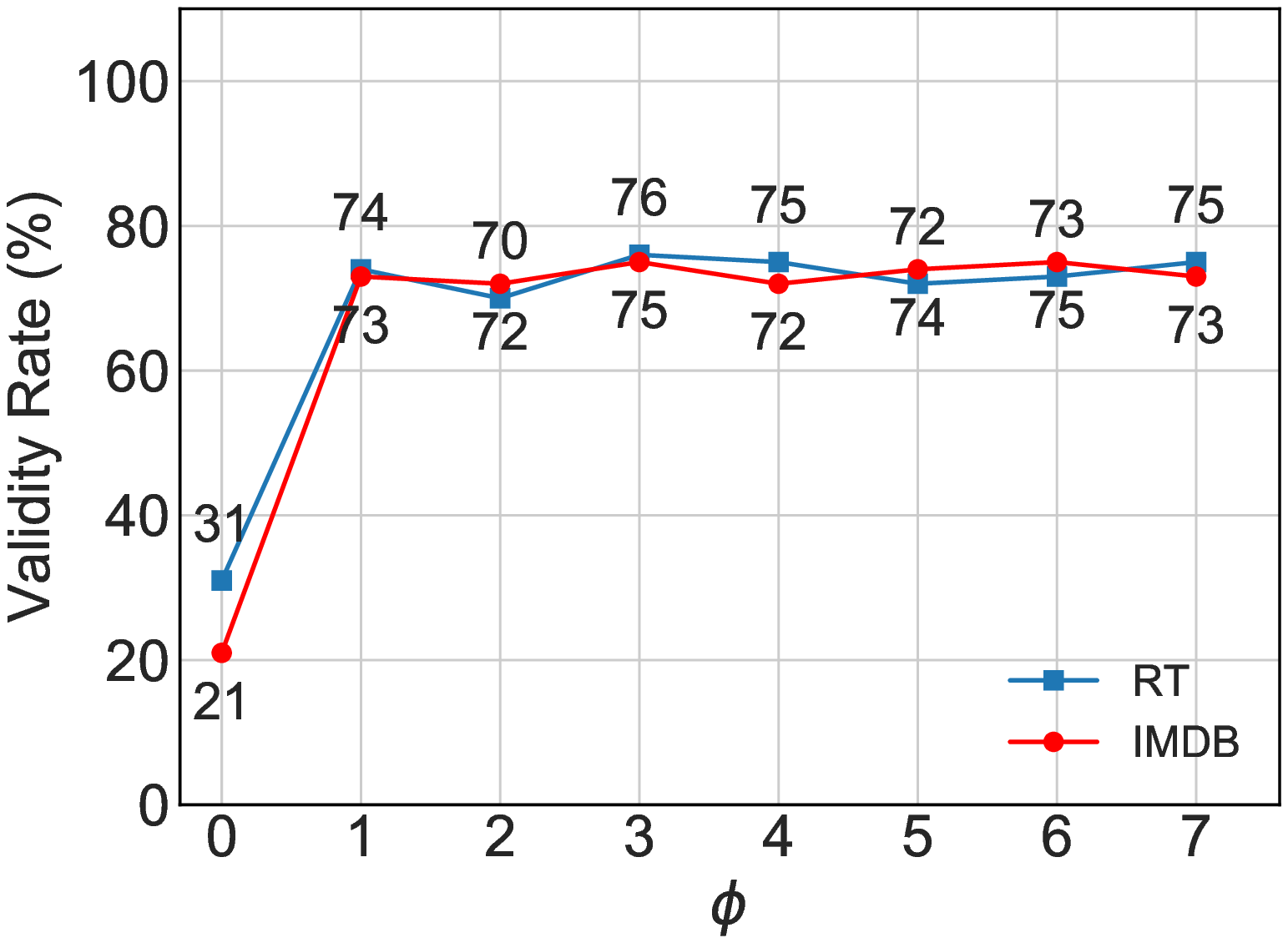}
}%
\caption{The attack success rate, perplexity and validity of unrestricted adversarial text generation from scratch. Randomly sample $z$ to generate adversarial texts from scratch with different $\phi$. Note that when $\phi=0$, the model is a vanilla VAE}
\label{fig:random_gen}
\end{figure*}

We summarize the performances of of our method and all baselines in Table \ref{table:trans_success}. From Table \ref{table:trans_success}, we can observe that randomly changing $10\%$ words is not enough to fool the classifier. 
This implies the difficulty of attack. 
TextBugger and our method both achieve quite high attack success rate. While our method performs even better than TextBugger, which is the state-of-the-art method.

We show some adversarial examples generated by our method and TextBugger to demonstrate the differences in Figure \ref{fig:trans_examples}. 

We can observe that TextBugger mainly changes the spelling of words. The generated text becomes not fluent and easy to be detected by grammar checking systems. Also,  though humans may guess the original meanings, the changed words are treated as out of vocabulary words by models. For example, TextBugger changes the spelling of ‘awful’, ‘cliches’ and ‘foolish’ in Figure \ref{fig:trans_examples}. These words are important negative sentiment words for a negative sentence. It is natural that changing these words to unknown words can change the prediction of models. Unlike TextBugger, our method generates meaningful and fluent contents. For example, in the first example of Figure \ref{fig:trans_examples}, we replace ‘read the novel’ with ‘love the book’, the substitution is still fluent and make sense to both humans and models. 

\smallskip
\noindent
\textbf{Generation Speed.} It takes about one hour and about 3 hours to train our model on RT dataset and IMDB dataset respectively. We also  evaluate the time cost of generating one adversarial example. We take the FGSM method as the representative of gradient-based  methods, as FGSM is the fastest among them. We measure the time cost of generating $1,000$ adversarial examples and calculate the average time of generating one. Results are shown in Table \ref{table:gen_time}. 

\begin{table}
\caption{Time cost of generating one adversarial text.}
\label{table:gen_time}
\centering
\begin{tabular}{l|c|c|c}
\toprule
\textbf{Method} & FGSM+NNS & TextBugger & Ours ($\phi = 5$) \\ \midrule
\textbf{Time}            & 0.7s              & 0.05s               & \textbf{0.014s}  
\\ \bottomrule
\end{tabular}
\end{table}

We can observe that our method is much faster than others. That is mainly because our generative model is trained beforehand. After the model is trained, the generation of one batch just requires one feed-forward. 

\subsection{Unrestricted Adversarial Text Generation} \label{section:random_gen}

\begin{figure*}[!t]
\centering
\includegraphics[width=0.9\textwidth, angle=0]{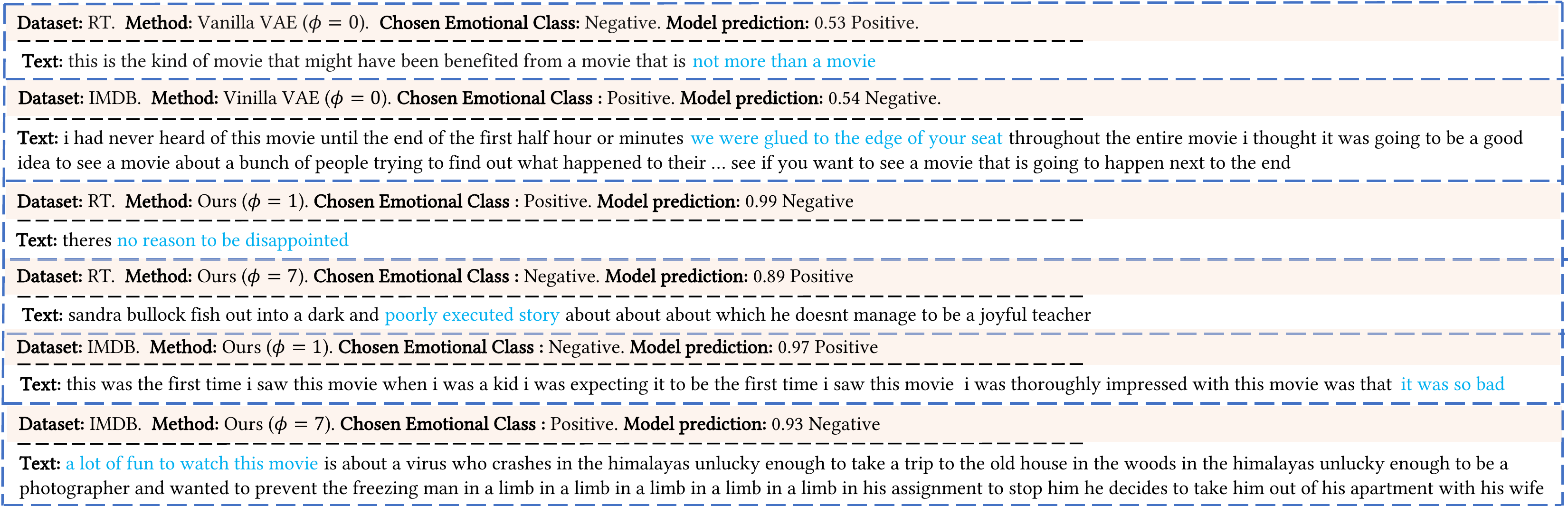}
\caption{Adversarial examples generated from scratch unrestrictedly. Humans should classify adversarial texts as the chosen emotional class $y_k$.}
\label{fig:random_gen_examples}
\end{figure*}

As mentioned in Section \ref{section:text_generator}, after our model is trained, we can randomly sample $z$ from latent space, choose a desired class $y_k \in \mathcal{Y}$ , get the embedding vector $c_k$ of $y_k$,  then feed $[z, c_k]$ to the decoder to generate  adversarial texts unrestrictedly with no need of labeled text. 

\smallskip
\noindent
\textbf{Attack Success Rate.}
When training, we can tune $\phi$ in Equation  \ref{eq:joint_loss} to affect the model. After trained with different $\phi$, we observe the generated texts are different. We  randomly generate 50,000 examples and compute the proportion of adversarial examples with different $\phi$. The results are shown in Figure \ref{fig:random_gen.suc_rate}. Notice if we set $\phi = 0$, the model is a vanilla VAE and  it is not trained continually after  pretrained.

From Figure \ref{fig:random_gen.suc_rate}, we can observe that the attack success rate of the vanilla VAE is only $10.3\%$ and $20.1\%$ respectively, this implies that only randomly generating texts can hardly fool the targeted model. When $\phi$ is greater than $0$, the attack success rate is consistently better than the vanilla VAE. This reflects the importance of $\mathcal{L}_{adv}$. 

Also, the attack success rate increases as $\phi$ becomes larger. 
It is because the larger $\phi$ is, the more important role $\mathcal{L}_{adv}$ will plays in the final joint loss $\mathcal{L}_{joint}$. So, the text generator $\mathcal{G}$ is more easily guided by the $\mathcal{L}_{adv}$ to generate an adversarial example. 

To evaluate the quality of the generated adversarial texts with different $\phi$,  we adopt three metrics : perplexity, validity and diversity.

\begin{figure*}[!t]
\centering
\subfigure[RT]{
\label{fig:data_aug.rt}
\includegraphics[width=0.28\textwidth]{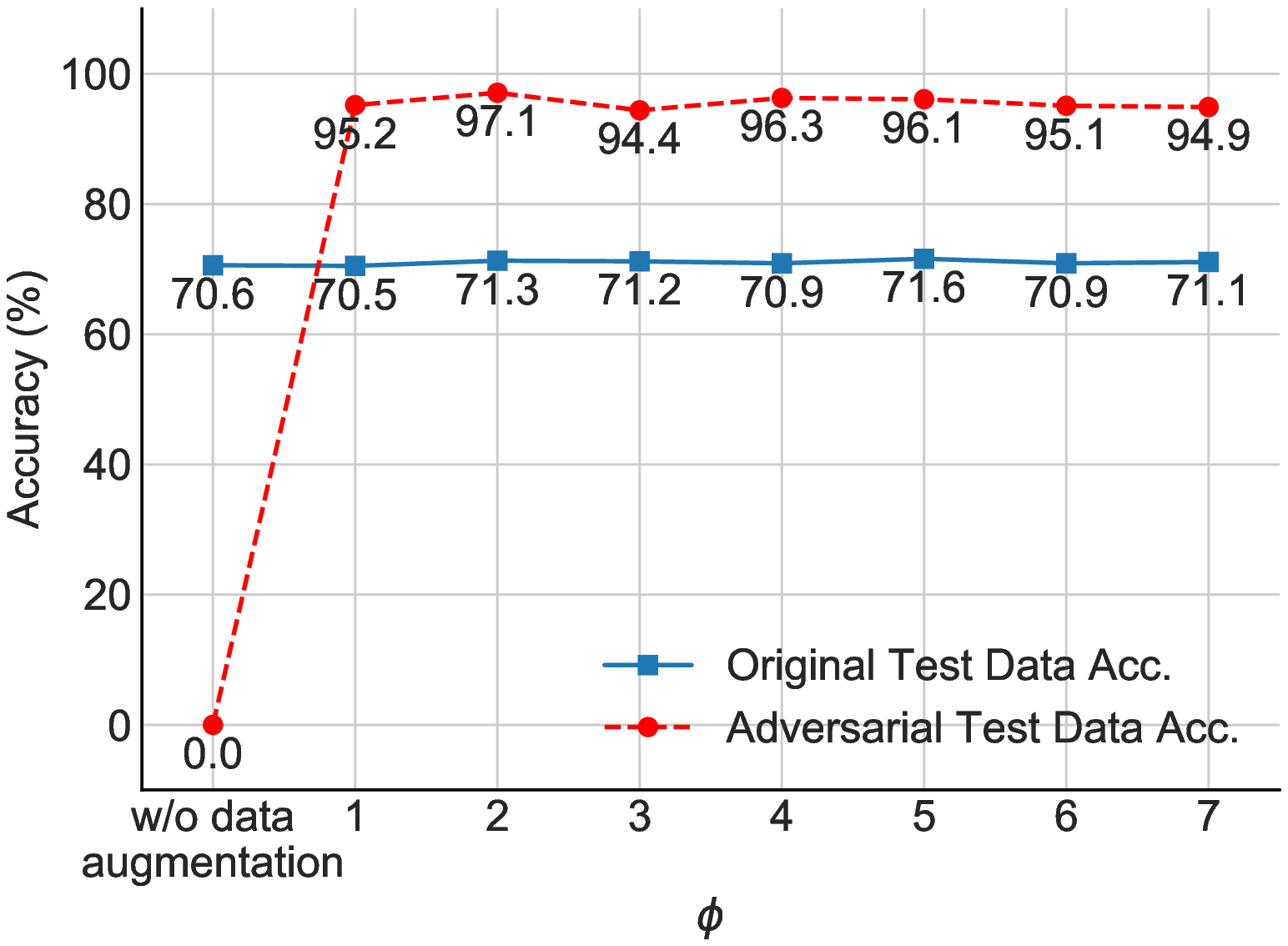}
}
\subfigure[IMDB]{
\label{fig:data_aug.IMDB}
\includegraphics[width=0.28\textwidth]{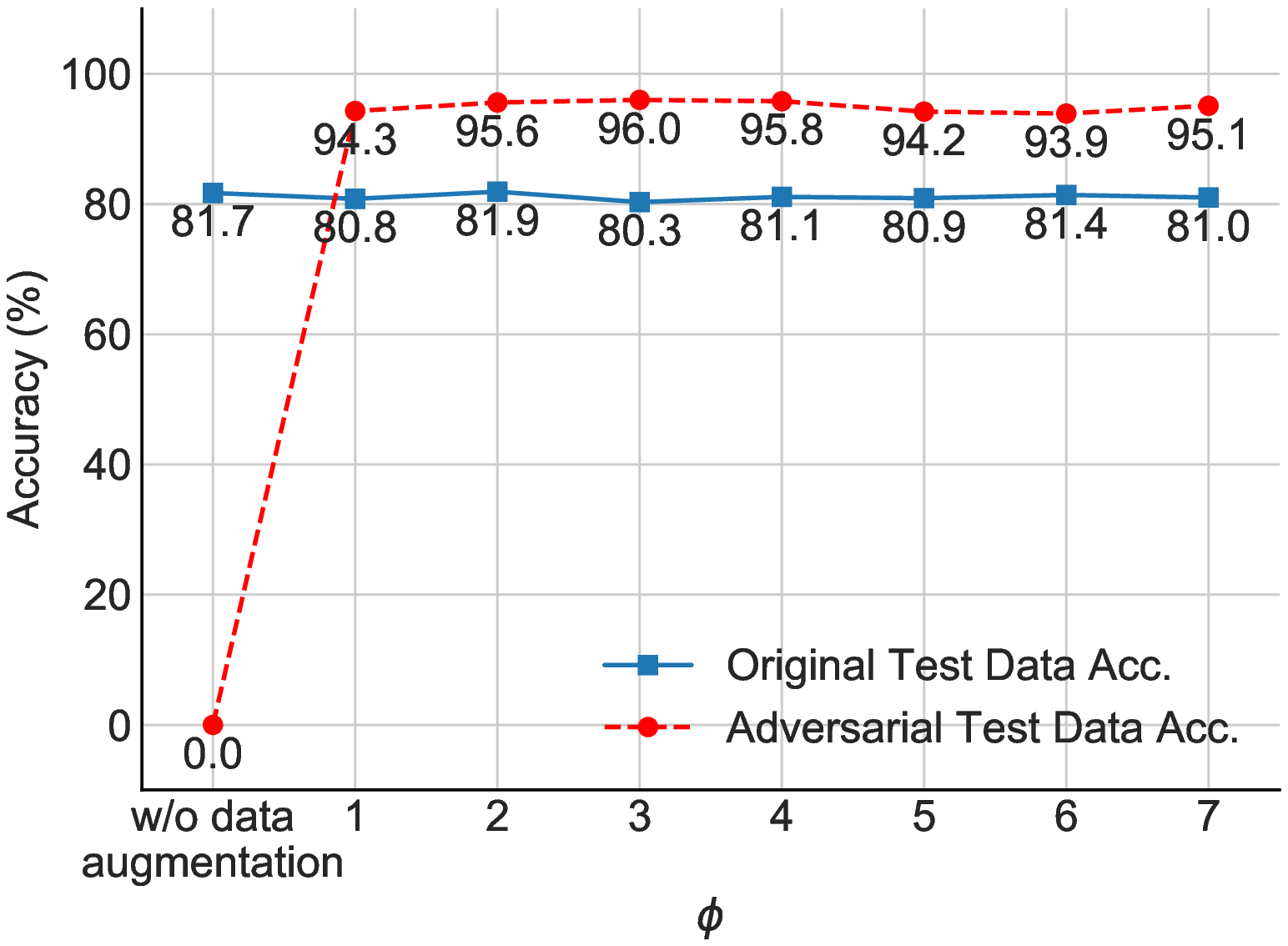}
}%
\subfigure[Data augmentation compare]{
\label{fig:data_aug.comp}
\includegraphics[width=0.28\textwidth]{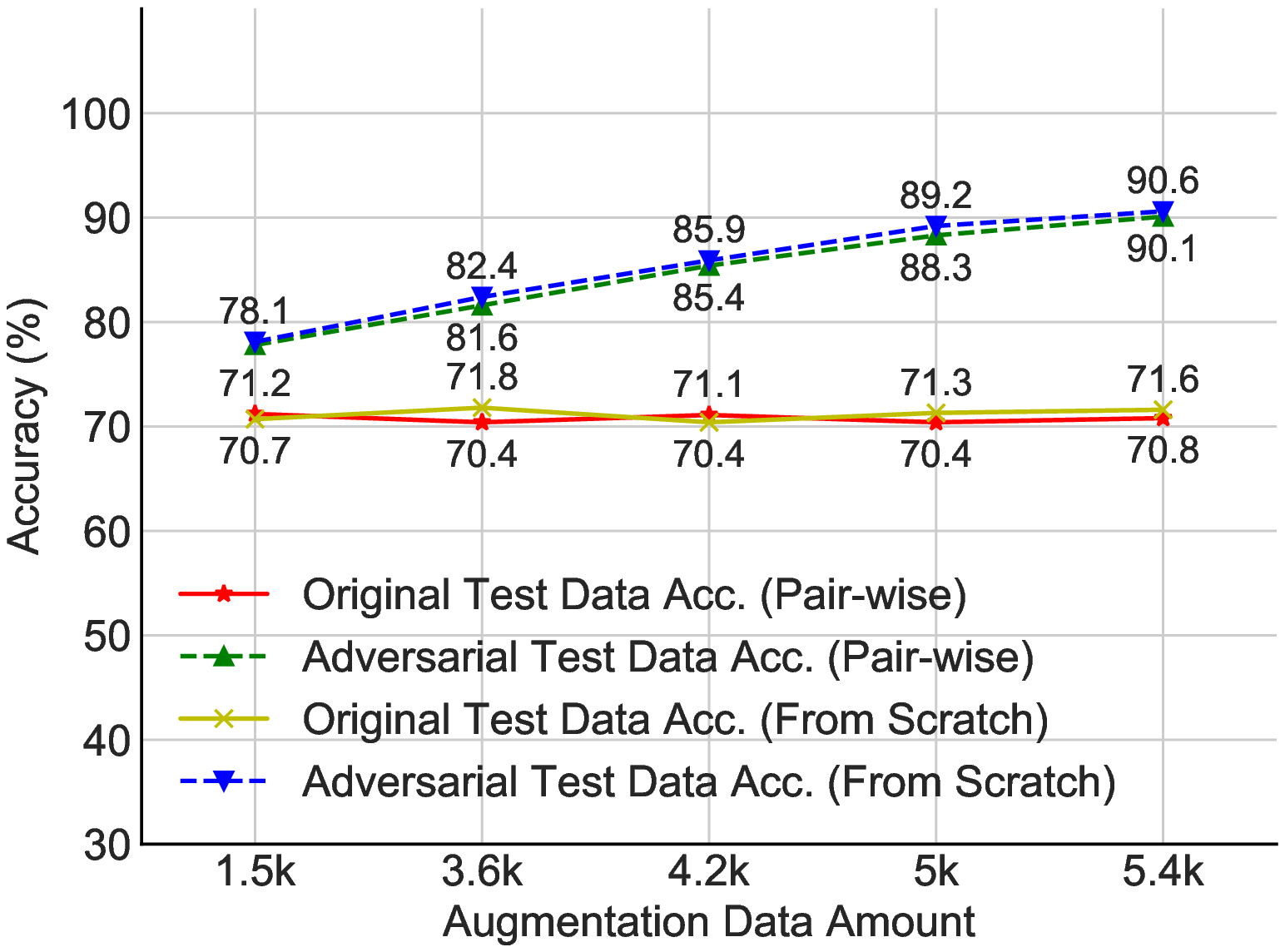}
}%
\caption{Defense with adversarial training in different settings. (a) and (b) On RT and IMDB datasets, data augmentation with adversarial data generated from scratch under different $\phi$. (c) On RT dataset, accuracy of models trained with equal size of augmentation adversarial data, which is generated in pair-wise way and unrestricted generation way respectively.}
\label{fig:data_aug}
\end{figure*}

\smallskip
\noindent
\textbf{Perplexity.}
Perplexity \cite{brown1992estimate} is a measurement of how well a probability model predicts a sample. A low perplexity indicates the language model is good at predicting the sample. Given a pretrained language model, it can also be used to evaluate the quality of texts. Similarly, a low perplexity indicates the text is more fluent for the language model. We compute perplexity as:
\begin{equation}
    Perplexity = - \frac{1}{|word\_num|}\sum_{x \in {X}'}\sum^{V}_{j=1}\log P({x}'_j|{x}'_0,...,{x}'_{j-1}) 
    \label{eq:perplexity}
\end{equation}
where $V$ is the number of words in one sentence. $P({x}'_j)$ is the probability of $j$-th word in ${x}'$ computed by the language model.

We train a language model with the training data of IMDB and RT, and use it as $P$ in Equation \ref{eq:perplexity}. We measure and compare the perplexity of the generated 50,000 texts and data of the original training set. Results are shown in Figure \ref{fig:random_gen.perplexity}.
We can observe that the perplexity is only a bit higher than the original data's, which means that the quality of generated texts are acceptable. Also, as $\phi$ gets larger, the perplexity gets bigger. This is perhaps because $\mathcal{L}_{adv}$ can distort the generated texts.

\smallskip
\noindent
\textbf{Validity.}
If we feed $[z, c_k]$ to the decoder, then a valid generated adversarial text is supposed to be classified as class $y_k$ by humans but be classified as class $y_t \neq y_k$ by the targeted model. 
We randomly select 100 generated texts for each $\phi$ and manually evaluate their validity. The results are shown in Figure \ref{fig:random_gen.validity}.

From Figures \ref{fig:random_gen.validity}, we can observe that the validity rates of our method on both datasets are higher than $70\%$ and much higher than that of the vanilla VAE.
This implies our methods can generate high-quality and high-validity texts with high attack success rate.

\smallskip
\noindent
\textbf{Diversity.}
We first generate one million adversarial texts. To compare generated texts with train data, we extract all 4-grams of train data and generated texts. On average, for each generated text, less than $18\%$ of 4-grams can be found in all 4-grams of train data on all datasets. This shows that there exists some similarity and our model can also generate texts with different words combinations. To compare generated texts with each other, we suppose that if over $20\%$ of 4-grams of one generated text don't exist at the same time in any one of the other generated texts, the text is one unique text. We observe more than $70\%$ of generated texts are unique. This proved that the generated texts are diverse.

\smallskip
\noindent
\textbf{Adversarial Examples.}
We show some valid adversarial examples generated by our method in Figure  \ref{fig:random_gen_examples}. We can view that the adversarial examples generated by the vanilla VAE is more likely neutral, and the confidence of the targeted model is not huge. 
On the contrary, the generated examples of our method have high confidence of the targeted model. This shows $\mathcal{L}_{adv}$ is important to attack success rate. Besides, the fluency and validity of texts generated by our method are acceptable.

\subsection{Ablation Study} \label{section:ablation}

In this section, we further demonstrate the effectiveness of discrimators.
We now report the ablation study.

We first remove discrimators and $\mathcal{L}_{disc}$, then train our model. We compare it with the model trained with $\mathcal{L}_{joint}$ in a  min-max game. We evaluate their attack success rate, perplexity and validity. Results are show in Table \ref{table:ablation}. 

\begin{table}
\caption{Performance of our model trained with and without $\mathcal{L}_{disc}$.}
\label{table:ablation}
\centering
\scalebox{0.85}{
\begin{tabular}{c|c|ccc}
\toprule
\textbf{Dataset}      & \textbf{Method}              & \textbf{\begin{tabular}[c]{@{}c@{}}Attack Success\\ Rate\end{tabular}} & \textbf{Perplexity} & \textbf{Validity} \\ \toprule
\multirow{2}{*}{RT}   & with $\mathcal{L}_{disc}$    & $90.2\%$                                                               & 2.79                & $75\%$            \\ \cline{2-5} 
                      & without $\mathcal{L}_{disc}$ & $94.1\%$                                                               & 7.32                & $15\%$            \\ \toprule
\multirow{2}{*}{IMDB} & with $\mathcal{L}_{disc}$    & $93.9\%$                                                               & 2.88                & $73\%$            \\ \cline{2-5} 
                      & without $\mathcal{L}_{disc}$ & $94.3\%$                                                               & 7.41                & $12\%$            \\ \bottomrule
\end{tabular}
}
\end{table}

The attack success rates of models trained with and without $\mathcal{L}_{disc}$ are close. But the validity of the model trained without $\mathcal{L}_{disc}$ is much lower than that of the model with $\mathcal{L}_{filter}$. The reason of this phenomenon is as follows.  When training the generator $\mathcal{G}$ with only $\mathcal{L}_{VAE}$ and $\mathcal{L}_{adv}$, suppose we want to generate positive adversarial texts and  the targeted model must classify it as negative, the easiest way to achieve this goal is to change a few words in the generated text to negative words, such as "bad". But texts generated this way can not fool humans. If we add discrimators to draw distribution of adversarial texts close to the distribution of real data, this phenomenon can be controlled. 
This shows that discrimators and the min-max game $\mathop{min}\mathop{max}\mathcal{L}_{disc}^{k}$  can improve the validity greatly.

\subsection{Defense With Adversarial Training} \label{section:random_aug_sec}

Using the adversarial examples to augment the  training data can make models more robust, this is called adversarial training.

On RT dataset, we randomly generate  4k adversarial texts to augment the training data and  1k to test the model. On IMDB dataset, we randomly generate 10k, of which 8k for training and 2k for testing.  Results are shown in Figure \ref{fig:data_aug.rt} and Figure \ref{fig:data_aug.IMDB}. 

Through adversarial data augmentation, test accuracy on the original test data is stable. Also, the accuracy on the adversarial data is improved greatly (from $0$ to $>90\%$). It implies that adversarial training can make models more robust without hurting its effectiveness.

Then, on RT dataset, we first augment training data with adversarial examples generated   by pair-wise generation. The adversarial examples are generated through transforming training data. Note that we have 8k training data in RT dataset. When we set bigger $\phi$, the attack success rate is higher, so we can generate more adversarial examples in the pair-wise way. 
But with any $\phi$, unrestricted generation from scratch can result in infinite adversarial data. We compare the adversarial data augmentation performances of pair-wise and unrestricted generation from scratch.
We use the same number of  adversarial examples generated by the two modes, and hold out $20\%$ of generated data for testing. Results are shown in Figure \ref{fig:data_aug.comp}. 

We can see that with pair-wise generation, if training data is limited, we need to generate more adversarial examples to improve the adversarial test accuracy. Higher adversarial test accuracy requires higher $\phi$.
But higher $\phi$ results in bigger perplexity, which means low text quality. Differently, with unrestricted generation from scratch, we can generate infinite adversarial texts using very small $\phi$, with high fluency and similar adversarial test accuracy.
Thus, under similar adversarial test accuracy, the text fluency of pair-wise generation is worse than that of unrestricted generation from scratch. This indicates the advantage of our proposed method.

\section{Conclusion}
In this paper, we have proposed a scalable  method to generate adversarial texts from scratch attacking a text classification model. We add an  adversarial  loss to enforce the generated text to mislead the targeted model. Besides, we use discrimators and GAN-like training strategy to make adversarial texts mimic real data of the desired class. After the generator is trained, it can generate diverse adversarial examples of a desired class on a large scale without real-world texts.  Experiments  show that the proposed method is scalable and can achieve higher attack success rate at a higher speed compared with recent methods. In addition, it is also demonstrated that the generated texts are of good quality and mostly valid. We further conduct ablation experiments to verify effects of discrimators. Experiments of data augmentation indicate that our method generates more diverse adversarial texts with higher quality than pair-wise generation, which can make the targeted model more robust.


\bibliographystyle{ecai}
\bibliography{ecai}
\end{document}